\documentclass{ecai} 

\usepackage{latexsym}
\usepackage{amssymb}
\usepackage{amsmath}
\usepackage{amsthm}
\usepackage{booktabs}
\usepackage{enumitem}
\usepackage{graphicx}
\usepackage{color}

\newcommand{\BibTeX}{B\kern-.05em{\sc i\kern-.025em b}\kern-.08em\TeX}

\begin{document}


\begin{frontmatter}


\paperid{123} 


\title{Multi-Objective Reinforcement Learning for Large Language Model Optimization: Visionary Perspective}


\author[A]{\fnms{Lingxiao}~\snm{Kong}\thanks{Corresponding Author. Email: lingxiao.kong@fit.fraunhofer.de}}
\author[B]{\fnms{Cong}~\snm{Yang}}
\author[A,C]{\fnms{Oya}~\snm{Deniz Beyan}}
\author[A,C]{\fnms{Zeyd}~\snm{Boukhers}}

\address[A]{Fraunhofer Institute for Applied Information Technology FIT, Germany}
\address[B]{Soochow University, China}
\address[C]{University Hospital of Cologne, Germany}

\begin{abstract}
Multi-Objective Reinforcement Learning (MORL) presents significant challenges and opportunities for optimizing multiple objectives in Large Language Models (LLMs). We introduce a MORL taxonomy and examine the advantages and limitations of various MORL methods when applied to LLM optimization, identifying the need for efficient and flexible approaches that accommodate personalization functionality and inherent complexities in LLMs and RL. We propose a vision for a MORL benchmarking framework that addresses the effects of different methods on diverse objective relationships. As future research directions, we focus on meta-policy MORL development that can improve efficiency and flexibility through its bi-level learning paradigm, highlighting key research questions and potential solutions for improving LLM performance.
\end{abstract}

\end{frontmatter}


\section{Introduction} 
In Large Language Models (LLMs), various generation traits are learned and optimized, such as reflection and fluency in response generation~\citep{do2022pair}. However, many LLM tasks contain such predefined objectives but use them only as evaluation metrics rather than incorporating them into the LLM to enhance performance on these metrics. Multi-Objective Reinforcement Learning (MORL) can address this gap by learning RL policies that explicitly focus on multiple objectives~\cite{hayes2022practical}. Applying MORL presents several key challenges: balancing objectives while maintaining overall performance, approximating Pareto-optimality, achieving computational efficiency during training and inference, ensuring stability, adapting to diverse preferences, scaling to accommodate new objectives, and providing explainability regarding objective contributions. Furthermore, LLMs involve large-scale parameters and complex sequential generation, while RL encounters computational complexity and overoptimization issues, making current methods \textbf{inflexible} and \textbf{inefficient}~\cite{lin2024reinforcement}.

MORL typically uses scalar-based RL to define explicit, decomposed objectives for LLM optimization through automatic metrics or AI feedback~\cite{lee2025rlaif}. It can directly optimize explicit objectives without requiring extensive human effort and addresses the limitation that humans cannot effectively quantify objectives, while improving scalability to new preferences and objectives. Effectively incorporating explicit competing objectives within LLMs can further support Reinforcement Learning with Human Feedback (RLHF)~\cite{ouyang2022training}, leading to improved solutions for LLMs in real-world applications. Our research reveals particular promise for meta-policy MORL methods, which can address the inefficiency and inflexibility challenges inherent in conventional MORL approaches, making them well-suited for LLM optimization~\cite{feriani2022multiobjective}. However, meta-policy methods remain underexplored in mainstream MORL literature, while many MORL methods remain limited in their application to decision-making tasks. This gap underscores the need for comprehensive benchmarking of current MORL methods for LLMs, utilizing suitable evaluation metrics and focusing on analyzing objective relationships. Such benchmarking analysis would provide clearer guidance for developing more robust MORL approaches in the future.

This paper presents a MORL taxonomy with analysis of advantages and limitations, and proposes a vision for systematic benchmarking of these methods on LLM tasks. We also highlight key research questions and potential solutions to address the limitations of current meta-policy methods in future work.

\section{Visionary Perspective} 
\label{sec:vision}

\subsection{MORL Taxonomy}

\begin{figure}[t!]
\begin{center}
\includegraphics[width=1.0\linewidth]{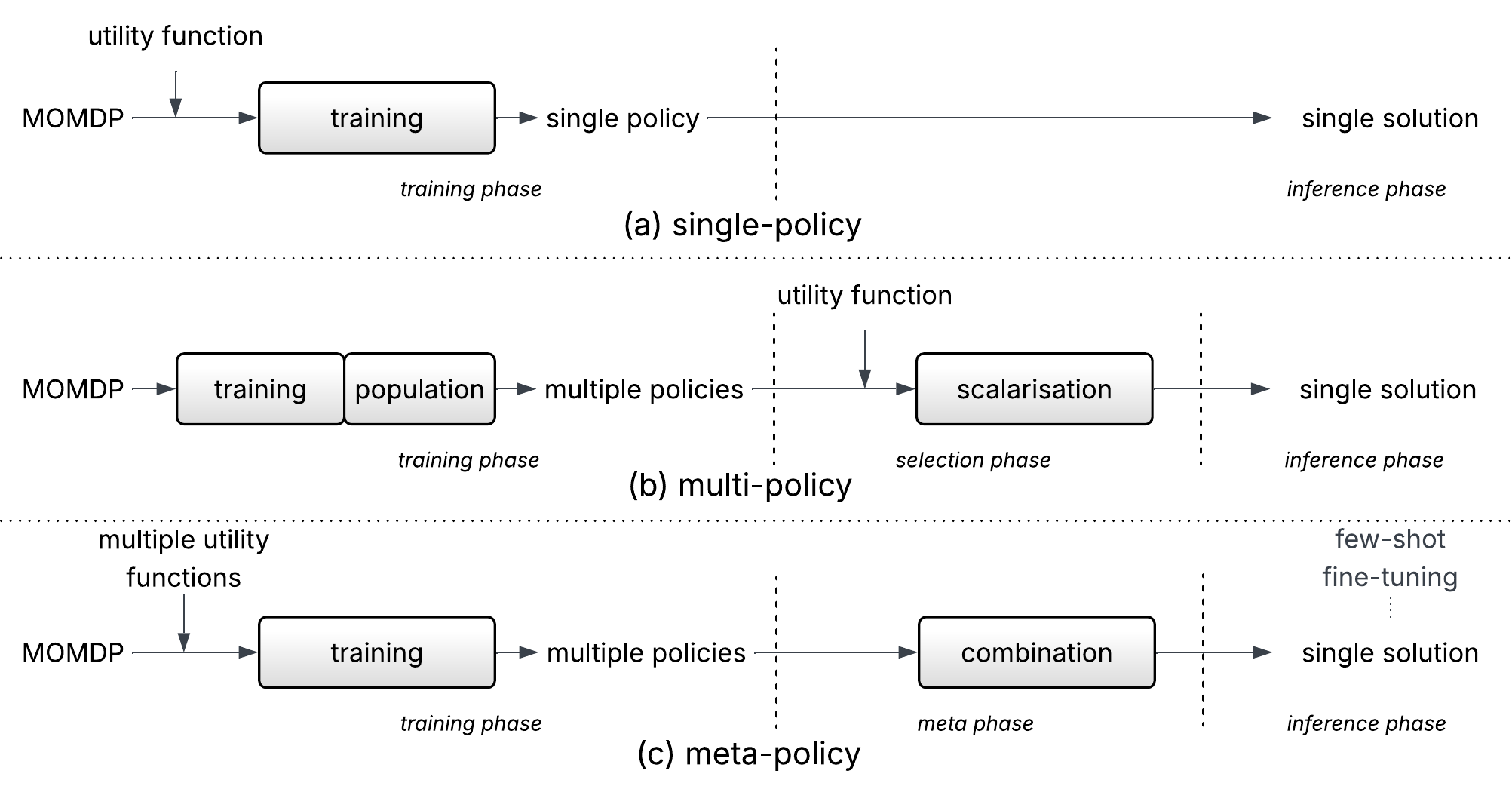}
\end{center}
\vspace{-2em}
\caption{MORL methods demonstration. }
\label{fig:method_demo}
\end{figure}

Many works categorize mainstream MORL methods into only single-policy (with prior knowledge) and multi-policy (with posterior knowledge) approaches~\citep{luukkonen2023artificial,felten2023toolkit}. However, \cite{feriani2022multiobjective} first introduced meta-policy methods, which incorporate meta-learning and ensemble learning strategies~\cite{ye2024multi, song2024ensemble}, into the MORL taxonomy. We have also conducted preliminary work that applies these strategies to develop an efficient and flexible meta-policy approach with enhanced performance~\citep{kong2025emorl}. Figure~\ref{fig:method_demo} illustrates the fundamental workflows of each MORL method category, from Multi-Objective Markov Decision Process (MOMDP) formulation to final solution~\cite{lizotte2016multi}.

\textbf{Single-policy} methods are the most conventional and require prior knowledge to define utility functions for user-specific objective preferences. While straightforward to train, they suffer from key limitations: scalarized reward weights do not correspond linearly to performance, and adapting to different utility functions requires retraining from scratch. Dynamic weighting techniques using Multi-Armed Bandit and Markov Decision Tabular mechanisms have been proposed to address these issues~\citep{perez2024dynamic,pu2024dynamic}. However, they still struggle with diverse objective preferences and increase computational costs without guaranteed improvements ~\citep{kong2025emorl}.

\textbf{Multi-policy} methods use posterior knowledge to apply utility functions to MORL models~\citep{hayes2022practical}. They employ MOMDP concepts and separate rewards into vectors. These methods learn numerous policies with varying performance across reward dimensions and utilize mechanisms such as evolutionary algorithms to identify Pareto-optimal solutions and populate new policies~\citep{callaghan2025extending}. Posterior knowledge then serves as a utility function to select the single optimal solution matching user-specific preferences from the Pareto-optimal set. However, for large-scale LLMs and complex RL fine-tuning algorithms, training numerous policies and computing population mechanisms becomes computationally inefficient~\citep{liang2024evolutionary}.

\textbf{Meta-policy} methods primarily use meta-learning and ensemble learning to combine trained policies into meta-policies~\citep{ye2024multi,song2024ensemble}. These strategies provide adaptability and scalability, and have been proven effective in many domains for enhancing model robustness and transfer learning capabilities~\citep{malik2024attention}. Meta-policy methods typically employ \textbf{bi-level learning}: training several policies at the lower level and combining them at the upper level. The combined policy can be further fine-tuned to adapt to different objective preferences with minimal steps~\citep{ye2024multi}. By training a few sampled policies for combination, meta-policy methods address the inflexibility and inefficiency of single-policy and multi-policy methods. However, they introduce new challenges: performance degradation on specific preferences while failing to approximate Pareto-optimality~\citep{chen2023efficient}.

\subsection{MORL Benchmarking}
To further develop these MORL methods, we need a systematic understanding of their effects. However, systematic benchmarking studies of MORL methods for LLM optimization are particularly lacking. The goal of our envisioned MORL benchmarking is to analyze the \textbf{relationship between objectives}, as competing objectives and correlated objectives necessitate different degrees of optimization~\cite{shah2016pareto}. For example, aggregating model parameters might work for correlated objectives but not for competing objectives~\cite{rame2023rewarded}. It is valuable to examine different LLM generation traits, identify their relationships, and evaluate various MORL methods using the following benchmarking metrics: 

1) \textbf{Performance} on user-specific objective preferences, measured as a pass ratio or scalar value by weighting objective scores with preferences. 2) \textbf{Pareto-optimality}, evaluating how outcome models are theoretically Pareto-optimal using hypervolume metrics~\citep{hayes2022practical}. 3) \textbf{Training efficiency}, measuring data and time consumption or computational resources during training under identical configurations (e.g., batch size and seed number). 4) \textbf{Inference efficiency}, assessing computational complexity during inference, measured by generation time, throughput, and latency. 5) \textbf{Stability}, conducting experiments with different seeds to measure performance and efficiency variance, indicating method reliability. 6) \textbf{Adaptability}, measuring the capability to adapt to different preferences through few-shot fine-tuning. 7) \textbf{Scalability}, qualitatively assessing the effort required to incorporate new preferences and objectives. 8) \textbf{Explainability}, determining how each objective contributes to performance for discovering precise user-specific preferences.

\subsection{Meta-Policy Development}
A crucial perspective in LLM tasks is that users have different preferences regarding trade-offs between objectives and might update their preferences over time~\cite{shao2023eliciting}. Thus, many metrics beyond performance and Pareto-optimality have become essential for acquiring multi-objective solutions in LLM optimization. Meta-policy methods represent a promising emerging approach that addresses these challenges. Among these methods, parameter aggregation like \cite{rame2023rewarded} suffers from alignment issues: combining parameters from different models may fail to effectively correspond to objective-specific features when objectives compete with each other. Logit aggregation like \cite{shi2024decoding} shows significant degradation on fluency objectives since at the logit level, much of the rich intermediate contextual information has been compressed into final predictions. Directly combining these logits makes output generation incoherent. Since LLMs handle sequential understanding, generation, and reasoning, maintaining connections between models is crucial compared to classification or regression tasks. In our preliminary work \cite{kong2025emorl}, we proposed training separate models for each objective, using hierarchical grid search to find optimal preference weights, and aggregating \textbf{hidden states} to preserve contextual features. This approach outperforms parameter-level and token-level aggregation with improved performance, efficiency, scalability, and explainability. However, its performance remains degraded compared to single-policy and multi-policy methods, raising directions for future research.

Building on the method in our preliminary work, we propose a valuable direction to develop \textbf{Mixture-of-Experts (MoE)} for addressing MORL, where expert networks are trained separately (at a lower level) and a gating network is then learned to effectively combine the expert outputs (at an upper level)~\cite{cai2025moe}. We identify two critical research questions and corresponding potential solutions for advancing the MoE-based methods at the lower and upper levels respectively: (1) \textbf{how to train individual models while preserving performance across objectives}, enhancing expert training at the lower level to ensure Pareto optimality using gradient update strategies, such as Multi-Gradient Descent Algorithm (MGDA)~\cite{desideri2012multiple}, and (2) \textbf{how to intelligently combine trained models to balance objectives with contexts and preferences}, refining expert combination at the upper-level to effectively integrate multiple objectives using dynamic weighting, such as Contextual Multi-Armed Bandits (CMABs)~\cite{bietti2021contextual}.

\section{Discussion} 
\label{sec:discussion}

This paper provides a visionary review of recent MORL research and identifies that applying MORL to LLM optimization presents unique challenges due to requiring personalization functionality and inherent complexities in LLMs and RL, yet receives limited attention in the literature. We propose establishing benchmarks for MORL methods in LLM optimization using comprehensive evaluation metrics to research the effects of methods on different objective relationships. Among existing MORL approaches, meta-policy methods show promise in addressing computational inefficiency and preference inflexibility, but they remain underexplored and exhibit performance limitations. We propose developing MoE-based methods with potential training and combination strategies that can improve LLM performance as a future research direction.







\bibliography{mybibfile}

\end{document}